\documentclass{article}

\usepackage[final]{neurips_2021}

\usepackage[utf8]{inputenc}
\usepackage[T1]{fontenc}
\usepackage{url}
\usepackage{booktabs}
\usepackage{amsfonts}
\usepackage{nicefrac}
\usepackage{microtype}
\usepackage[hidelinks]{hyperref} 
\usepackage[inline]{enumitem}
\usepackage[table, dvipsnames]{xcolor}
\usepackage{graphicx}
\usepackage{amsmath}
\usepackage{float}
\usepackage{subcaption}
\usepackage{adjustbox}
\usepackage{caption} 
\usepackage{mathtools, nccmath}
\usepackage{tikz}
\usetikzlibrary{shapes.geometric}
\usepackage{changepage}

\newcommand*{\tabindent}{\hspace{3mm}}
\newcommand{\colcircle}[1]{\tikz\node[circle,draw=#1, fill=#1, inner sep=1.75pt] (d) at (0,0) {};}
\newcommand{\coldiamond}[1]{\tikz\node[diamond,draw=#1, fill=#1, inner sep=1.25pt] (d) at (0,0) {};}
\newcommand{\colsquare}[1]{\tikz\node[draw=#1, fill=#1, minimum width=0.5pt, minimum height=0.5pt, inner sep=2.5pt] (d) at (0,0) {};}
\definecolor{darkblue}{HTML}{1A254B}
\definecolor{lightblue}{HTML}{A7BED3}
\definecolor{blue}{HTML}{114083}
\definecolor{green}{HTML}{81B5AE}
\definecolor{pink}{HTML}{F2545B}
\definecolor{red}{HTML}{A4243B}

\title{Multi-Scale Representation Learning on Proteins}

\author{%
  Vignesh Ram Somnath\thanks{Equal contribution.}\\
  Dept. of Computer Science\\
  ETH Zurich \\
  \texttt{vsomnath@ethz.ch} \\
  \And
  Charlotte Bunne$^*$ \\
  Dept. of Computer Science\\
  ETH Zurich \\
  \texttt{bunnec@ethz.ch} \\
  \And
  Andreas Krause \\
  Dept. of Computer Science\\
  ETH Zurich \\
  \texttt{krausea@ethz.ch} 
}

\begin{document}

\maketitle

\begin{abstract}
    Proteins are fundamental biological entities mediating key roles in cellular function and disease. This paper introduces a multi-scale graph construction of a protein -- \textsc{HoloProt} -- connecting surface to structure and sequence. The surface captures coarser details of the protein, while sequence as primary component and structure -- comprising secondary and tertiary components -- capture finer details. Our graph encoder then learns a multi-scale representation by allowing each level to integrate the encoding from level(s) below with the graph at that level. We test the learned representation on different tasks, \begin{enumerate*}[label=(\roman*.)]
    \item ligand binding affinity (\emph{regression}), and
    \item protein function prediction (\emph{classification})
    \end{enumerate*}.
    On the regression task, contrary to previous methods, our model performs consistently and reliably across different dataset splits, outperforming all baselines on most splits. On the classification task, it achieves a performance close to the top-performing model while using 10x fewer parameters. To improve the memory efficiency of our construction, we segment the multiplex protein surface manifold into \emph{molecular} superpixels and substitute the surface with these superpixels at little to no performance loss.
\end{abstract}

\section{Introduction}

Protein design and engineering has become a crucial component of pharmaceutical research and development, finding application in a wide variety of diagnostic and industrial settings.
Besides understanding the design principles determining structure and function of proteins, current efforts seek to further enhance or discover proteins with properties useful for technological or therapeutic applications.
To efficiently guide the search in the vast design space of functional proteins, we need to be able to robustly predict properties of a candidate protein \citep{yang2019}. 
Moreover, understanding role and function of proteins is crucial to study causes and mechanism of human disease \citep{fessenden2017}.

To achieve this, representations incorporating the complex nature of proteins are required.
Proteins consist of amino acids, organic molecules linked by peptide bonds forming a linear \emph{sequence}. Each of the twenty amino acids carries a unique side chain, giving rise to an incomprehensibly large combinatorial space of possible protein sequences.
The primary sequence drives the folding of polymers -- a spontaneous process guided by hydrophobic interactions, formation of intramolecular hydrogen bonds, and van der Waals forces into a unique three-dimensional \emph{structure}. The resulting shape and \emph{surface} manifold with rich physiochemical properties carry essential information for understanding function and potential molecular interactions.

\begin{figure*}[t]
    \centering
    \includegraphics[width=\textwidth]{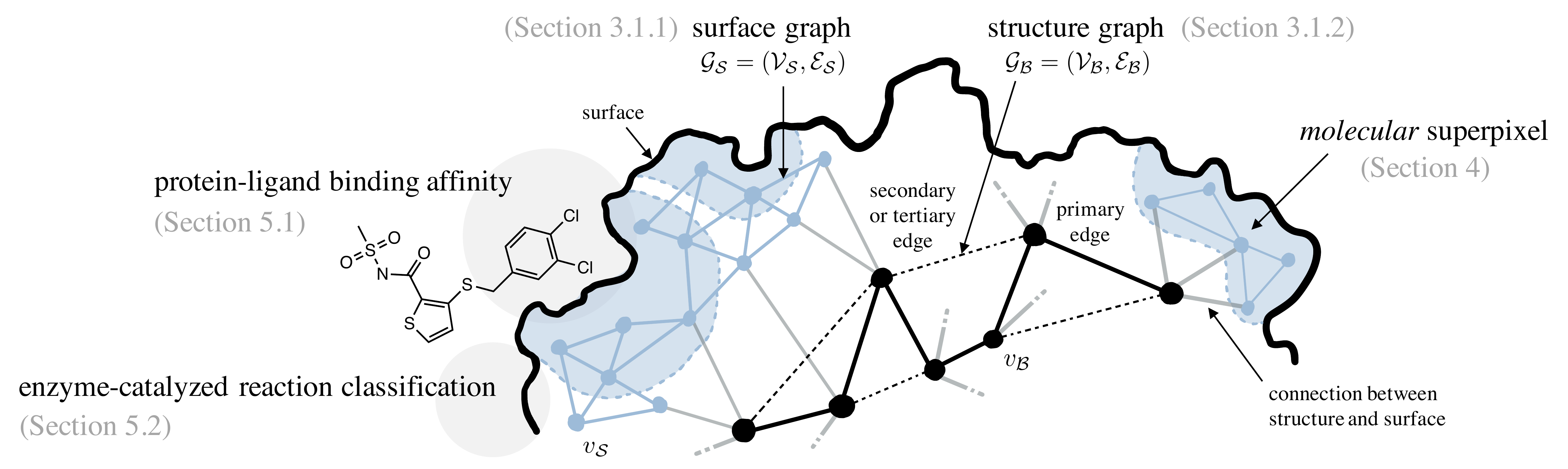}
    \caption{\textbf{Overview of \textsc{HoloProt}} Our multi-scale protein representation algorithm integrates primary, secondary and tertiary elements of protein structures and connects them to the surface. We extract higher-level protein motifs by introducing \emph{molecular} superpixels. Both structure and surface are represented as graphs $\mathcal{G}_\mathcal{B}$ and $\mathcal{G}_\mathcal{S}$, respectively. The method is evaluated on two representative tasks, protein-ligand binding affinity and enzyme-catalyzed reaction classification.}
    \label{fig:overview}
\end{figure*}

Previous methods typically only consider an individual subset within these scales, focusing on either sequence \citep{ozturk2018, hou2018}, three-dimensional structure \citep{hermosilla2021, derevyanko2018} or surface \citep{gainza2020}.
Two proteins with similar sequences can fold into entirely different conformations. While these proteins might catalyze the same type of reactions, their behavior to specific inhibiting drugs might be divergent. Interaction between proteins and ligands, on the other hand, is controlled by molecular surface contacts \citep{gainza2020}. Molecular surfaces, determined by subjacent amino acids, are fingerprinted with patterns of geometric and chemical properties, and thus their integration in protein representations is crucial.

In this work, we present a novel multi-scale graph representation which integrates and connects the complex nature of proteins across all levels of information. \textsc{HoloProt} consists of a surface and structure layer (both represented as graphs) with explicit edges between the layers. Our construction is guided by the intuition that propagating information from surface to structure would allow each residue to learn encodings reflective of not just its immediate residue neighborhood, but also the higher-level geometric and chemical properties that arise from interactions between a residue and its neighborhood. The associated multi-scale encoder then learns representations by integrating the encoding from the layer below, with the graph at that layer (Section~\ref{sec:ms_prot}). Such multi-scale representations have been previously used in molecular graph generation \citep{jin2020hierarchical} with impressive results.

We further improve the memory efficiency of our construction by segmenting the large and rich protein surface into \emph{molecular ``superpixels''}, summarizing higher-level fingerprint features and motifs of proteins. Substituting the surface layer with these superpixels results in little to no performance degradation across the evaluated tasks. The concept of \emph{molecular superpixels} might be of interest beyond our model (Section~\ref{sec:superpixel}). 

The multi-objective and multi-task nature of protein engineering poses a challenge for current methods, often designed and evaluated only on specific subtasks of protein design. By incorporating the biology of proteins, strong representations exhibit robust performance across tasks.
We demonstrate our model's versatility and range of applications by deploying it to tasks of rather distinct nature, including a regression task, e.g., inference of protein ligand binding affinity, and classification tasks, i.e., enzyme-catalyzed reaction classification (Section~\ref{sec:evaluation}).

\section{Related Work}

\paragraph{Protein Representation Learning}
With increasing availability of sequence and structure data, the field of protein representation learning has advanced rapidly, with methods falling largely in one of the following categories:

\begin{itemize}[wide=0pt, leftmargin=15pt, label=]

\item \emph{Sequence-based methods.}
One-dimensional amino acid sequences continue to be the simplest, most abundant source of protein data and various methods have been developed that borrow architectures developed in natural language processing (NLP). One-dimensional convolutional neural networks have been used to classify a protein sequence into folds and enzyme function \citep{hou2018, dalkiran2018}, and to predict their binding affinity to ligands \citep{ozturk2018}.  
Furthermore, methods have applied complex NLP models trained unsupervised on millions of unlabeled protein sequences and fine-tuned them on different downstream tasks \citep{rao2019, elnaggar2020, bepler2019}. Despite being advantageous when only the sequence is available, these methods ignore the full spatial complexity of proteins. 

\item \emph{Structure-based methods.}
To learn beyond sequences, approaches have been developed, that consider the 3D structure of proteins. 
3D convolutional neural networks have been utilized for protein quality assessment \citep{derevyanko2018}, protein contact prediction \citep{townshend2019} and protein-ligand binding affinity tasks \citep{ragoza2017protein, jimenez2018, townshend2020}. 
An alternate representation treats proteins as graphs, applying graph neural networks for enzyme classification \citep{dobson2005}, interface prediction \citep{fout2017}, and protein structure quality prediction \citep{baldassarre2021}. \citet{gligorijevic2021} use a long short term memory cell (LSTM) to encode the sequence, followed by a graph convolutional network (GCN) \citep{kipf2017} to capture the tertiary structure, and apply this to the function prediction task.
\citet{hermosilla2021} propose a convolutional operator that learns to adapt filters based on the primary, secondary, and tertiary structure of a protein, showing strong performance on reaction and fold class prediction.

\item \emph{Surface-based methods.} Taking a different viewpoint, \citet{gainza2020} hypothesize that the protein surface displays patterns of chemical and geometric features that
fingerprint a protein's interaction with other biomolecules. They utilize geodesic convolutions, which are extensions of convolutions on surfaces, and learn fingerprint vectors, showing improved performance across binding pocket and protein interface prediction tasks. 
\end{itemize}

\paragraph{Protein Motif Detection}
Protein motifs have largely been synonymous with common and conserved patterns in a protein's sequence or structure influencing protein function, e.g., the helix-turn-helix motif binds DNA.
Understanding these fragments is essential for 3D structure prediction, modeling, and drug design.
While reliably detecting evolutionary motifs, existing tools \citep{golovin2008} do not provide a full segmentation of the protein surface manifold. Our work takes a different viewpoint, by looking at protein motifs from the context of a protein surface. Previous methods developed in this context either only consider geometric information rather than physiological properties \citep{cantoni2010}, are computationally expensive \citep{cantoni2011}, or designed for particular downstream tasks \citep{stepniewska2020}. Our molecular superpixel approach provides a task-independent segmentation utilizing both geometric and chemical features, while also being computationally efficient.

\section{Multi-Scale Protein Representation} \label{sec:ms_prot}
In this section, we describe our multi-scale graph construction and the associated encoder. Figure~\ref{fig:overview} illustrates the main principles of \textsc{HoloProt}. We represent a protein $\mathcal{P}$ as a graph $\mathcal{G}_{\mathcal{P}}$ with two layers capturing different scales:

\begin{enumerate}[label=(\roman*.)]
    \item \textbf{Surface layer.} This layer captures the coarser representation details of a protein. The protein surface is generated using the triangulation software MSMS \citep{connolly1983, sanner1996}. We represent this layer as a graph $\mathcal{G}_\mathcal{S}$, where each surface node $u_\mathcal{S}$ has a feature vector $\mathbf{f}_{u_\mathcal{S}}$ denoting its charge, hydrophobicity and local curvature \citep{gainza2020}. Two surface nodes $(u_\mathcal{S}, v_\mathcal{S})$ have an edge if they are part of a triangulation. Each surface node additionally has a residue identifier $r$, indicating the amino acid residue it corresponds to. Multiple surface nodes can have the same residue identifier.
    
    \item \textbf{Structure layer.} This layer captures the finer representation details of a protein. A protein typically has four structural levels: (i.) primary structure (sequence), (ii.) secondary structure ($\alpha$-helices and $\beta$-sheets), (iii.) tertiary structure (3D structure) and (iv.) quaternary structure (complexes) \citep{fout2017}. We represent this layer as a graph $\mathcal{G}_\mathcal{B}$, where each node $u_\mathcal{B}$ corresponds to a residue $r$. Two nodes $(u_{\mathcal{B}}, v_{\mathcal{B}})$ have an edge in $\mathcal{G}_\mathcal{B}$ if the C$_{\alpha}$ atoms of the two nodes occur within a certain distance of each other. Distance based thresholding ensures that different structural levels are implicitly captured in the neighborhood of a node $u_\mathcal{B}$.
\end{enumerate}

We further introduce edges from the surface layer to the structure layer in order to propagate information between them. Specifically, we introduce a directed edge between a surface node $u_\mathcal{S}$ and a backbone node $u_\mathcal{B}$ if they both have the same residue identifier $r$. Typically, we have between 20-40 surface nodes $\{u_{\mathcal{S}}\}$ that map to the same structure node $u_\mathcal{B}$. This gives us the multi-scale graph which is then encoded by our multi-scale message passing network.
Details on the features used for both the structure and surface layer can be found in Appendix~\ref{app:features}.

\subsection{Multi-Scale Encoder}
Our multi-scale message passing network uses one {\em message passing neural network (MPN)} for each layer in the multi-scale graph \citep{lei2017, gilmer2017}. This allows us to learn structured representations of each scale, which can then be tied together through connections between the scales. Before detailing the remainder of the architecture, we introduce some notational preliminaries. For simplicity, we denote the MPN encoding process as $\text{MPN}_{\theta}(\cdot)$ with parameters $\theta$. We denote $\text{MLP}_{\theta}(\mathbf{x}, \mathbf{y})$ for a multi-layer perceptron (MLP) with parameters $\theta$, whose input is the concatenation of $\mathbf{x}$ and $\mathbf{y}$, and $\text{MLP}_{\theta}(\mathbf{x})$ when the input is only $\mathbf{x}$. We also denote the residue identifier of a node $u$ with $\mathrm{id}(u)$, and the neighbors of a node $u$ as $\mathcal{N}(u)$. The details of the MPN architecture are listed in the Appendix~\ref{app:mpn}.

\subsubsection{Surface Message Passing Network}

We first encode the surface layer $\mathcal{G}_\mathcal{S}$ of the multi-scale protein graph $\mathcal{G}_{\mathcal{P}}$. The inputs to the MPN are node features $\mathbf{f}_{u_\mathcal{S}}$ and edge features $\mathbf{f}_{{u_\mathcal{S}}{v_\mathcal{S}}}$ of $\mathcal{G}_{\mathcal{S}}$. For more details on the input features used for surface nodes and edges, refer to Appendix~\ref{app:features}. 
The MPN (with parameters $\theta_\mathcal{S}$) propagates messages between the nodes for $K$ iterations, and outputs a representation $h_{u_\mathcal{S}}$ for each surface node $u_\mathcal{S}$,

\begin{equation*}
    \label{eqn:sur_mpn}
    \{\mathbf{h}_{u_\mathcal{S}}\} =  \mathrm{MPN}_{\theta_{\mathcal{S}}}(\mathcal{G}_\mathcal{S}, \{\mathbf{f}_{u_\mathcal{S}}\}, \{\mathbf{f}_{{u_\mathcal{S}}{v_\mathcal{S}}}\}_{v_\mathcal{S} \in \mathcal{N}(u_\mathcal{S})}).
\end{equation*}

\subsubsection{Structure Message Passing Network}
For each node $u_\mathcal{B}$ in the structure layer $\mathcal{G}_{\mathcal{B}}$, we first prepare the input to the MPN (with parameters $\theta_{\mathcal{B}}$) by using an MLP (with parameters $\theta$) on the concatenated version of its initial features $\mathbf{f}_{u_\mathcal{B}}$ and the mean of the surface node vectors with the same residue identifier $S = \{\mathbf{h}_{u_\mathcal{S}} | \mathrm{id}(u_\mathcal{S}) = \mathrm{id}(u_\mathcal{B})\}$

\begin{equation*}
    \label{eqn:struct_feat}
    \mathbf{x}_{u_{\mathcal{B}}} = \mathrm{MLP}_{\theta}(\mathbf{f}_{u_\mathcal{B}}, \nicefrac{\sum_{S}{\mathbf{h}_{u_\mathcal{S}}}}{|S|}).
\end{equation*}

Given the edge features $\mathbf{f}_{{u_\mathcal{B}}{v_\mathcal{B}}}$, we then run $K$ iterations of message passing, to compute the representations $\mathbf{h}_{u_\mathcal{B}}$  for each structure node $u_\mathcal{B}$,

\begin{equation*}
    \label{eqn:struct_mpn}
    \{\mathbf{h}_{u_\mathcal{B}}\} =  \mathrm{MPN}_{\theta_{\mathcal{B}}}(\mathcal{G}_\mathcal{B}, \{\mathbf{x}_{u_\mathcal{B}}\}, \{\mathbf{f}_{{u_\mathcal{B}}{v_\mathcal{B}}}\}_{v_\mathcal{B} \in \mathcal{N}(u_\mathcal{B})}).
\end{equation*}

The graph representation $\mathbf{c}_{\mathcal{G}_\mathcal{P}}$ is an aggregation of structure node representations, 
\begin{equation}
    \label{eqn:graph_repr}
    \mathbf{c}_{\mathcal{G}_\mathcal{P}} = \sum_{u_\mathcal{B} \in \mathcal{G}_\mathcal{B}}{\mathbf{h}_{u_\mathcal{B}}}.
\end{equation}

\subsection{Task Specific Training}
This multi-scale encoding allows us to learn a structured representation of a protein tying different scales together, which can then be utilized for any downstream task. In this work, we evaluate our method on two rather distinct tasks \begin{enumerate*}[label=(\roman*.)]
    \item protein-ligand binding affinity regression, and
    \item enzyme-catalyzed reaction classification
\end{enumerate*}.
The architectural details for both downstream tasks are described below. These modules can be adapted and modified in order to utilize \textsc{HoloProt} for other use cases.

\subsubsection{Protein-Ligand Binding Affinity}
\label{sec:prot_lig_bind_aff}
Protein-ligand binding affinity prediction depends on the interaction of a protein, encoded using the \textsc{HoloProt} framework, and a corresponding ligand, in most cases small molecules.
To encode the ligand represented as a graph $\mathcal{G}_\mathcal{L}$, we use another $\text{MPN}$ (with parameters $\theta_{\mathcal{L}}$) and aggregate its node representations to obtain a graph representation $c_{\mathcal{G}_\mathcal{L}}$. We concatenate the graph representations $\mathbf{c}_{\mathcal{G}_\mathcal{P}}$ (Equation~\ref{eqn:graph_repr}) of the protein and $c_{\mathcal{G}_\mathcal{L}}$ of the ligand, and use that as input to a $\text{MLP}$ (with parameters $\phi$) to obtain predictions,
\begin{equation}
    \label{eqn:regression}
    s_a = \mathrm{MLP}_\phi(c_{\mathcal{G}_\mathcal{P}}, c_{\mathcal{G}_\mathcal{L}}).
\end{equation}

The model is trained by minimizing the mean squared error.

\subsubsection{Enzyme-Catalyzed Reaction Classification}
\label{sec:prot_func_pred}

To predict the enzyme-catalyzed reaction class, we use the graph representation $\mathbf{c}_{\mathcal{G}_\mathcal{P}}$ of the protein obtained via \textsc{HoloProt} as the input to a $\text{MLP}$ (with parameters $\phi$) to obtain the prediction logits,

\begin{equation}
    \label{eqn:classification}
    p_k = \mathrm{MLP}_\phi(c_\mathcal{G}).
\end{equation}
The model is trained by minimizing the cross-entropy loss.


\section{Superpixels on Molecular Surfaces}\label{sec:superpixel}
\begin{figure}
    \centering
    \includegraphics[width=.99\textwidth]{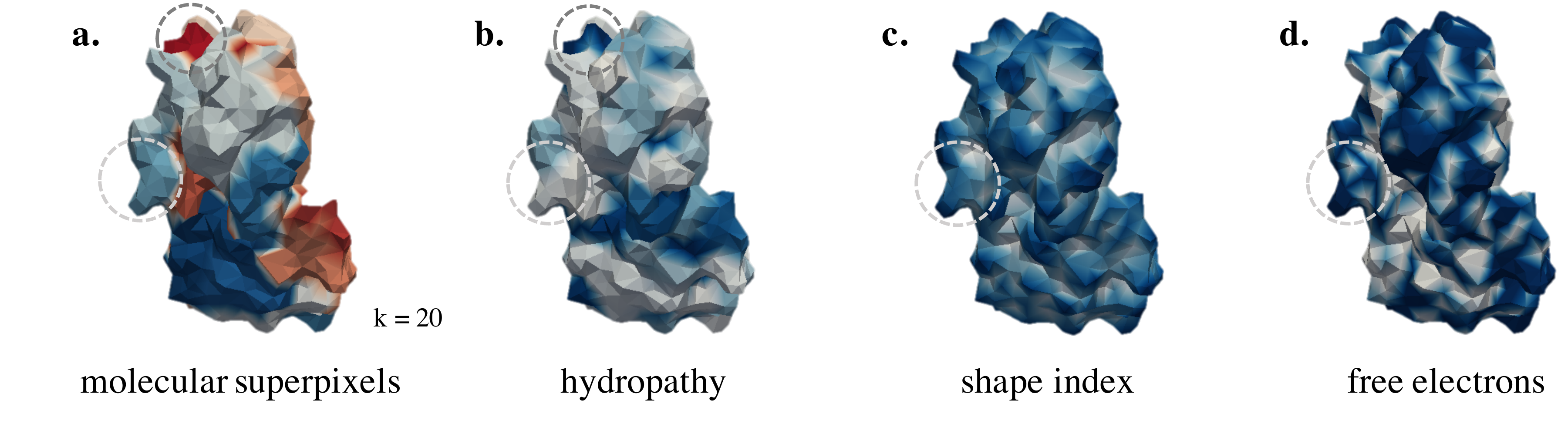}
    \caption{\looseness -1 \textbf{Molecular Superpixels and Surface Features of the HIV-1 Protease (\texttt{PDB ID: 2AVQ}).} \textbf{a.}
    Molecular superpixels, indicated by different colors ($k = 20$), and the corresponding surface features, i.e.,  \textbf{b.} hydropathy, \textbf{c.}  shape index, and \textbf{d.} free electrons. As highlighted, molecular superpixels are spatially compact and overlap with surface regions dominated by single features such as hydrophobic patches while capturing coherent areas across all surface features. The protein complex contains 198 residues.}
    \label{fig:superpixel_2avq}
\end{figure}

Protein surface manifolds are complex and represented via large meshes. In order to improve the computational and memory efficiency of our construction, we introduce the notion of \emph{molecular superpixels}. Originally developed in computer vision \citep{ren2003, mori2004, kohli2009}, superpixels are defined as perceptually uniform regions in the image. In the molecular context, we refer to superpixels as segments on the protein surface capturing higher-level fingerprint features and protein motifs such as hydrophobic binding sites.

In order to apply the segmentation principle to three-dimensional molecular surfaces, we employ graph-based superpixel algorithms on triangulated surface meshes.
The superpixel representation of the protein surface needs to satisfy several requirements, as
\begin{enumerate*}[label=(\roman*.)]
    \item molecular superpixels should not reduce the overall achievable performance of \textsc{HoloProt}, and
    \item molecular superpixels need to form geometrically compact clusters, and overlap with surface regions that are coherent in physiological surface properties, e.g., capture hydrophobic binding sides or highly charged areas.
\end{enumerate*}
Popular graph-based segmentation tools such as \citet[FH]{felzenszwalb2004}, mean shift \citep{comaniciu2002}, and watershed \citep{vincent1991}, however, produce non-compact superpixels of irregular sizes and shapes.
By posing the segmentation task as a maximization problem on a graph maximizing over 
\begin{enumerate*}[label=({\color{pink}\roman*.})]
    \item the entropy rate of the random walk on the surface graph $\mathcal{G}_{\mathcal{S}} = (\mathcal{V}_{\mathcal{S}}, \mathcal{E}_{\mathcal{S}})$ favoring the formation of compact and homogeneous clusters, and
    \item a balancing term encouraging clusters with similar sizes,
    \end{enumerate*}
the entropy rate superpixel (ERS) segmentation algorithm \citep{liu2011} outperforms previous methods across different tasks \citep{stutz2018} and achieves the desired properties of \emph{molecular} superpixels.

In order to incorporate geometric and chemical features of the surface $\mathbf{F}_\mathcal{S}$, we extend the surface graph  $\mathcal{G}_{\mathcal{S}} = (\mathcal{V}_{\mathcal{S}}, \mathcal{E}_{\mathcal{S}})$ with a non-negative similarity measure $w$, given as 
$w_{ij} = \sum_{\mathbf{f} \in \mathbf{F}_\mathcal{S}} |\mathbf{f}_{v_i} \mathbf{f}_{v_j}|$ for nodes $v_i$ and $v_j$ if connected by an edge $e_{ij}$.
We simulate a random walk $\mathbf{X} = \{X_t | t \in T, X_t \in \mathcal{V}_\mathcal{S}\}$ on a protein surface mesh, where the transition probability $p_{ij}$ between two nodes $v_i$ and $v_j$ is defined as $p_{ij} = P(X_{t+1} = v_j |X_t = v_i) = \nicefrac{w_{ij}}{w_i}$, where $w_i = \sum_{k : e_{ik} \in \mathcal{E}_\mathcal{S}} w_{ik}$.The corresponding stationary distributions of nodes $\mathcal{V}_\mathcal{S}$ are given by
\begin{align*}
    \boldsymbol{\mu}=\left(\mu_{1}, \mu_{2}, \ldots, \mu_{|\mathcal{V}_\mathcal{S}|}\right)^{\top}=\left(\frac{w_{1}}{w_{T}}, \frac{w_{2}}{w_{T}}, \ldots, \frac{w_{|\mathcal{V}_\mathcal{S}|}}{w_{T}}\right)^{\top}.
\end{align*}
Molecular superpixels are then defined by a subset of edges $\mathcal{M} \subseteq \mathcal{E}_\mathcal{S}$ such that the resulting graph, $\mathcal{G}_\mathcal{S} = (\mathcal{V}_\mathcal{S}, \mathcal{M})$, contains exactly $k$ connected
subgraphs.
Computing \emph{molecular} superpixels is achieved via optimizing the objective function with respect to the edge set $\mathcal{M}$
\begin{align*}
    & \max_{\mathcal{M}} - \underbrace{\sum_{i} \mu_{i} \sum_{j} p_{ij}(\mathcal{M}) \log \left(p_{ij}(\mathcal{\mathcal{M}})\right)}_{\textstyle \text{({\color{pink} i.}) entropy rate} \mathstrut}
    - \underbrace{\sum_{i} p_{Z_{\mathcal{M}}}(i) \log \left(p_{Z_{\mathcal{M}}}(i)\right) - n_{\mathcal{M}}}_{\textstyle \text{({\color{pink} ii.}) balancing function} \mathstrut} \\
    &\text{s.t. } \mathcal{M} \subseteq \mathcal{E}_{\mathcal{S}} \text{ and } n_\mathcal{M} \ge k, 
\end{align*}
where $n_\mathcal{M}$ is the number of connected components in the graph, $p_{Z_\mathcal{M}}$ denotes the distribution of cluster memberships $Z_\mathcal{M}$, and $\lambda \ge 0$ is the weight of the balancing term. 
Both terms satisfy monotonicity and submodularity and can thus be efficiently optimized based on techniques from submodular optimization \citep{nemhauser1978}. For further details on the entropy rate superpixel algorithm, see \citet{liu2011}.

A molecular superpixel $m$ comprising $k$ surface vertices is then given as $\mathbf{f}_m = (\mathbf{f}_{v_1}, \dots, \mathbf{f}_{v_k})$ for all $\mathbf{f} \in \mathbf{F}_\mathcal{S}$.
We summarize the feature representation of each molecular superpixel via the graph $\mathcal{G}_\mathcal{M} = (\mathcal{V}_\mathcal{M}, \mathcal{E}_\mathcal{M})$, where each node $m \in \mathcal{V}_\mathcal{M}$ is represented via (\texttt{mean}($\mathbf{f}_m$), \texttt{std}($\mathbf{f}_m$), \texttt{max}($\mathbf{f}_m$), \texttt{min}($\mathbf{f}_m$)) for all $\mathbf{f} \in \mathbf{F}_\mathcal{S}$ and an edge $e \in \mathcal{E}_\mathcal{M}$ via the Wasserstein distance between neighboring superpixels.

Figure~\ref{fig:superpixel_2avq} demonstrates \emph{molecular} superpixels for the enzyme HIV-1 protease \citep{brik2003}.
Besides being spatially compact, superpixels overlap with surface regions dominated by single features such as hydrophobic patches, while capturing coherent areas across all surface features.
Further examples of superpixels are displayed in Appendix~\ref{app:ex_superpixel}.

\section{Evaluation} \label{sec:evaluation}

Successful protein engineering requires optimization of multiple objectives. When searching for a protein with desired functionality, auxiliary but crucial properties such as stability measured in terms of free energy of folding also need to be satisfied. Furthermore, the field is also subject to a plethora of potential tasks and applications. 
In order to capture the multi-objective and multi-task nature of protein engineering,
we evaluate our method on two representative tasks: regression of the binding affinity between proteins and their ligands, and classification of enzyme proteins based on the type of reaction they catalyze. 

\subsection{Protein-Ligand Binding Affinity Prediction}\label{sec:pdbbind}
Studying the interaction between proteins and small molecules is crucial for many downstream tasks, e.g., accelerating virtual screening for potential candidates in drug discovery or protein design to improve the output of an enzyme-catalyzed reaction.
The architecture of the regression module is described in Equation~\ref{eqn:regression}.
    
\paragraph{Dataset.}
The \textsc{PDBbind} database (version 2019) \citep{liu2017} is a collection of the experimentally measured binding affinity data for all types of biomolecular complexes deposited in the Protein Data Bank \citep{berman2000}.
After quality filtering for resolution and surface construction, the refined subset comprises a total of $4,709$ biomolecular complexes.
The binding affinity provided in \textsc{PDBbind} is experimentally determined and expressed in molar units of the inhibition constant ($K_i$) or dissociation constant ($K_d$). Similar to previous methods \citep{ozturk2018, townshend2020}, we do not distinguish both constants and predict negative log-transformed binding affinity $pK_d / pK_i$.
We split the dataset into training, test and validation splits based on the scaffolds of the corresponding ligands (\emph{scaffold}), or a 30\% and a 60\% sequence identity threshold (\emph{identity 30\%}, \emph{identity 60\%}) to limit homologous ligands or proteins appearing in both train and test sets. 

\paragraph{Baselines.}
For evaluating the overall performance on the regression task, we compare \textsc{HoloProt} against several baselines including current state-of-the-art methods on both tasks. This comprises sequence-based methods \citep{ozturk2018, rao2019, bepler2019, elnaggar2020} as well as methods based on the three-dimensional structure of proteins \citep{townshend2020, hermosilla2021}, and recent methods using  geometric deep learning on protein molecular surfaces \citep{gainza2020}.

\paragraph{Evaluation metrics.} 
For evaluating different methods, we use three metrics -- root mean squared error (RMSE), Pearson correlation coefficient, and Spearman correlation coefficient. We also include the mean and standard deviation across 3 experimental runs.

\paragraph{Results.}
\begin{table}[t]
\caption[Ligand Binding Affinity Prediction Results]{\textbf{Protein-Ligand Binding Affinity Prediction Results} Comparison predictive performance of ligand binding affinity using the PDBbind dataset \citep{liu2017} of \textsc{HoloProt} against other methods. Results are reported for 3 experimental runs.}
    \label{tab:res_pdbbind}
    \begin{adjustbox}{max width=1.05\linewidth}
    \begin{tabular}{lcccccccccc}
    \toprule
    \textbf{Model} & \textbf{\# Params} & \multicolumn{3}{c}{\textbf{Sequence Identity (30 \%)}} & &  \multicolumn{3}{c}{\textbf{Sequence Identity (60 \%)}} \\ \\
    
    & & RMSE & Pearson & Spearman & & RMSE & Pearson & Spearman \\
    \cmidrule{3-5}\cmidrule{7-9}
    \textbf{Sequence-based Methods} \\
    \tabindent \citet{ozturk2018} & 1.93 M & 1.866 $\pm$ 0.080 & 0.472 $\pm$ 0.022 & 0.471 $\pm$ 0.024 && 1.762 $\pm$ 0.261 & 0.666 $\pm$ 0.012 & 0.663 $\pm$ 0.015\\
    \tabindent \citet{bepler2019} & 48.8 M & 1.985 $\pm$ 0.006	& 0.165 $\pm$ 0.006	& 0.152 $\pm$ 0.024 && 1.891 $\pm$ 0.004 & 0.249 $\pm$ 0.006 & 0.275 $\pm$ 0.008 \\
    \tabindent \citet{rao2019} & 93.0 M & 1.890 $\pm$ 0.035 & 0.338 $\pm$ 0.044 & 0.286 $\pm$ 0.124 && 1.633 $\pm$ 0.016 & 0.568 $\pm$ 0.033 & 0.571 $\pm$ 0.021 \\
    \tabindent \citet{elnaggar2020} & 2.4M\footnotemark[1] & 1.544 $\pm$ 0.015 & 0.438 $\pm$ 0.053 & 0.434 $\pm$ 0.058 && 1.641 $\pm$ 0.016 & 0.595 $\pm$ 0.014 & 0.588 $\pm$ 0.009 \\
    \textbf{Surface-based Methods} \\
    \tabindent \citet{gainza2020} & 0.62 M & 1.484 $\pm$ 0.018 & 0.467 $\pm$ 0.020 & 0.455 $\pm$ 0.014 && 1.426 $\pm$ 0.017 & 0.709 $\pm$ 0.008 & 0.701 $\pm$ 0.011 \\
    \textbf{Structure-based Methods} \\
    \tabindent \citet{townshend2020}\footnotemark[2] & - & \textbf{1.429} $\pm$ \textbf{0.042} & 0.541 $\pm$ 0.029 & 0.532 $\pm$ 0.033 && 1.450 $\pm$ 0.024 & 0.716 $\pm$ 0.008 & 0.714 $\pm$ 0.009 \\
    \tabindent \citet{townshend2020}\footnotemark[3] & - & 1.936 $\pm$ 0.120 & \textbf{0.581} $\pm$ \textbf{0.039} & \textbf{0.647} $\pm$ \textbf{0.071} && 1.493 $\pm$ 0.010 & 0.669 $\pm$ 0.013 & 0.691 $\pm$ 0.010 \\
    \tabindent \citet{hermosilla2021} & 5.80 M & 1.554 $\pm$ 0.016 & 0.414 $\pm$ 0.053 & 0.428 $\pm$ 0.032 && 1.473 $\pm$ 0.024 & 0.667 $\pm$ 0.011 & 0.675 $\pm$ 0.019 \\
    \\
    \textsc{HoloProt} (\colcircle{lightblue}) & 1.44 M & 1.464 $\pm$ 0.006 & 0.509 $\pm$ 0.002 & 0.500 $\pm$ 0.005 && \textbf{1.365} $\pm$ \textbf{0.038} & \textbf{0.749} $\pm$ \textbf{0.014} & \textbf{0.742} $\pm$ \textbf{0.011} \\
    \textsc{HoloProt} (\coldiamond{darkblue}) & 1.76 M &  1.491 $\pm$ 0.004 & 0.491 $\pm$ 0.014 & 0.482 $\pm$ 0.017 && 1.416 $\pm$ 0.022 & 0.724 $\pm$ 0.011 & 0.715 $\pm$ 0.006 \\
    \bottomrule \\
    \end{tabular}
    \end{adjustbox}
    \vspace{-1.25em}
    \raggedright
    \begin{adjustbox}{max width=.67\linewidth}
    \begin{tabular}{lcccc}
    \toprule
    \textbf{Model} & \textbf{\# Params} & \multicolumn{3}{c}{\textbf{Scaffold}} \\ \\
    & & RMSE & Pearson & Spearman \\
    \cmidrule{3-5}
    \textbf{Sequence-based Methods} \\
    \tabindent \citet{ozturk2018} & 1.93 M & 1.908 $\pm$ 0.145 & 0.384 $\pm$ 0.014 & 0.387 $\pm$ 0.016\\
    \tabindent \citet{bepler2019} & 48.8 M & 1.864 $\pm$ 0.009 & 0.269 $\pm$ 0.002 & 0.285 $\pm$ 0.019\\
    \tabindent \citet{rao2019} & 93.0 M & 1.680 $\pm$ 0.055 & 0.487 $\pm$ 0.029 & 0.462 $\pm$ 0.051 \\
    \tabindent \citet{elnaggar2020} & 2.4M\footnotemark[1] & 1.592 $\pm$ 0.009 & 0.398 $\pm$ 0.027 & 0.409 $\pm$ 0.029 \\
    \textbf{Surface-based Methods} \\
    \tabindent \citet{gainza2020} & 0.62 M & 1.583 $\pm$ 0.132 & 0.416 $\pm$ 0.111 & 0.412 $\pm$ 0.126 \\
    \textbf{Structure-based Methods} \\
    \tabindent \citet{hermosilla2021} & 5.80 M & 1.592 $\pm$ 0.012 & 0.365 $\pm$ 0.024 & 0.373 $\pm$ 0.019 \\
    \\
    \textsc{HoloProt} (\colcircle{lightblue}) & 1.44 M & 1.523 $\pm$ 0.028 & 0.489 $\pm$ 0.019 & 0.491 $\pm$ 0.020\\
    \textsc{HoloProt} (\coldiamond{darkblue}) & 1.28 M & \textbf{1.516} $\pm$ \textbf{0.014} & \textbf{0.491} $\pm$ \textbf{0.016} & \textbf{0.493} $\pm$ \textbf{0.014} \\
    \bottomrule \\
    \multicolumn{4}{c}{\Large \colcircle{lightblue} \enspace full surface  \hfill \coldiamond{darkblue} \enspace molecular superpixels}
    \end{tabular}
    \end{adjustbox}
\end{table}
\footnotetext[1]{The embeddings obtained via \citet{elnaggar2020} were saved to disk, instead of finetuning the entire pretrained model.}
\footnotetext[2]{Equivariant neural network (ENN) on binding pocket only.}
\footnotetext[3]{Graph neural network (GNN) on binding pocket only.}

\looseness -1 Table~\ref{tab:res_pdbbind} displays the results on protein-ligand binding affinity.
\textsc{HoloProt} (\colcircle{lightblue}, \coldiamond{darkblue}) performs consistently well across different tasks and dataset splits, outperforming all methods on the splits \emph{scaffold} and \emph{identity 60\%}. 
On \emph{identity 30\%}, our method outperforms most baselines, while having lower variability across the evaluated metrics.
\textsc{HoloProt} with molecular superpixels (\coldiamond{darkblue}) performs similar to \textsc{HoloProt} on the entire surface, with no or little performance loss, suggesting that molecular superpixels capture meaningful biological motifs. We include the models from \citep{townshend2020} for completeness, but note that these models were trained only using the protein binding pocket. Binding sites on proteins are often structurally highly conserved regions \citep{panjkovich2010assessing}. Considering only binding pockets, which vary less between the train and test splits, provides an additional simplification making the task less challenging. All other baselines were tested on the full proteins. 

\subsection{Enzyme-Catalyzed Reaction Classification}\label{sec:enzyme}
Predicting the reaction class of enzymes without the use of sequence similarity allows for efficient screening of \emph{de novo} proteins, i.e., macromolecules without evolutionary homologs, for catalytic properties \citep{desjardins1997}.
The architecture of the classification module is described in Equation~\ref{eqn:classification}).

\paragraph{Dataset.}
Enzyme Commission (EC) numbers constitute an ontological system with the purpose of defining and organizing enzyme functions \citep{webb1992}. The four digits of an EC number are related in a functional hierarchy, where the first level annotates the main enzymatic classes, while the next levels constitute subclasses, e.g. the EC number of the HIV-1 protease is 3.4.23.16.  
This task aims at predicting the enzyme-catalyzed reaction class of a protein based on according to all four
levels of the EC number. We use the same dataset and splits as provided by \citep{hermosilla2021}, comprising $37, 428$ proteins from $384$ EC numbers, with $29,215$ instances for training, $2,562$ instances for validation, and $5,651$ for testing. For more details on dataset construction, we refer to \citet[Appendix C]{hermosilla2021}. 

\paragraph{Baselines.}
For the classification task, we again compare \textsc{HoloProt} against several baselines including sequence-based methods \citep{hou2018}, methods partially pretrained on millions of sequences \citep{rao2019, bepler2019, elnaggar2020} as well as methods utilizing principles of geometric deep learning \citep{kipf2017, derevyanko2018, hermosilla2021}. The values for different baselines are taken from \citep{hermosilla2021}.

\paragraph{Evaluation metric.}
Model performance is measured via the mean accuracy score.

\paragraph{Results.}
\begin{table}[t]
\footnotesize
    \caption[Enzyme-Catalyzed Reaction Classification Results]{\textbf{Enzyme-Catalyzed Reaction Classification Results} Comparison of classification accuracy of \textsc{HoloProt} against other methods.}
    \label{tab:res_enzyme}
    \vspace{5pt}
    \centering
    \begin{adjustbox}{max width=\linewidth}
    \begin{tabular}{lcc}
    \toprule
    \textbf{Model} & \textbf{Parameters} & \textbf{Reaction Class} \\
    & & Accuracy  \\
    \cmidrule{1-3}
    \textbf{Sequence-based Methods} \\
    \tabindent \citet{hou2018} & 41.7 M & 70.9 \% \\
    \tabindent \citet{bepler2019} & 31.7 M & 66.7 \% \\
    \tabindent \citet{rao2019} (Transformer) & 38.4 M & 69.8 \% \\
    \tabindent \citet{elnaggar2020} & 420.0 M & 72.2 \% \\
    \textbf{Structure-based Methods} \\
    \tabindent \citet{kipf2017} & 1.0 M & 67.3 \% \\
    \tabindent \citet{derevyanko2018} & 6.0 M & 78.8 \% \\
    \tabindent \citet{hermosilla2021} & 9.8 M & \textbf{87.2} \% \\
    \\
    \textsc{HoloProt} (\colcircle{lightblue}) & 0.64 M & 77.8 \% \\
    \textsc{HoloProt} (\coldiamond{darkblue}) & 0.64 M & 78.9 \% \\
    \bottomrule \\
    \multicolumn{3}{c}{\colcircle{lightblue} \enspace full surface  \hfill \coldiamond{darkblue} \enspace molecular superpixels}
    \end{tabular}
    \end{adjustbox}
\end{table}

We report the results of enzyme-catalyzed reaction classification in Table~\ref{tab:res_enzyme}. While our method (\colcircle{lightblue}, \coldiamond{darkblue}) is unable to outperform the current state-of-the-art method \citep{hermosilla2021}, we achieve equivalent, if not better results to other methods at a fraction of the parameters used. 
Molecular superpixels also capture biologically meaningful protein surface motifs, as evidenced by a small increase in the overall classification performance.


\subsection{Ablation Studies}

\begin{table}[t]
\caption[Ablation Studies]{\textbf{Ablation Studies Results} Evaluation of architectural design choices of \textsc{HoloProt} by analyzing the performance of its individual components as well as feature summarization of molecular superpixels.}
    \label{tab:res_ablation}
    \centering
    \begin{adjustbox}{max width=.8\linewidth}
    \begin{tabular}{lccccc}
    \toprule
    & \multicolumn{3}{c}{\textbf{Ligand Binding Affinity}} & & \textbf{Enzyme Class} \\
    \textbf{Model} & \multicolumn{3}{c}{Sequence Identity (30 \%)} & & \\ \\
    & RMSE & Pearson & Spearman & & Accuracy \\
    \cmidrule{2-4}\cmidrule{6-6}
    
    Structure & 1.476 $\pm$ 0.027 & 0.51 $\pm$ 0.029 & 0.503 $\pm$ 0.027 && 74.2 \% \\
    Surface & 1.482 $\pm$ 0.015	& \textbf{0.512} $\pm$ \textbf{0.022} & \textbf{0.505} $\pm$ \textbf{0.017} && 28.6 \% \\
    \textsc{HoloProt} (\colcircle{lightblue}) & \textbf{1.464} $\pm$ \textbf{0.006} & 0.509 $\pm$ 0.002 & 0.500 $\pm$ 0.005 && 77.8 \% \\
    \textsc{HoloProt} (\coldiamond{darkblue}) &  1.491 $\pm$ 0.004 & 0.491 $\pm$ 0.014 & 0.482 $\pm$ 0.017 && \textbf{78.9} \% \\
    \textsc{HoloProt} (\colsquare{pink}) & 1.491 $\pm$ 0.027 & 0.503 $\pm$ 0.005 & 0.492 $\pm$ 0.004 && 75.7 \% \\
    \bottomrule \\
    \multicolumn{6}{c}{\colcircle{lightblue} \enspace full surface  \hfill \coldiamond{darkblue} \enspace molecular superpixels \hfill \colsquare{pink} \enspace molecular superpixel with MPN} 
    \end{tabular}
    \end{adjustbox}
\end{table}

To further evaluate the contribution of \textsc{HoloProt} to learning multi-scale protein representations, we conduct several ablation studies.
First, we analyze if the performance of the multi-scale model outperforms its isolated components, i.e. when using only structure or surface representation for subsequent downstream tasks.
The second ablation axis analyzes the construction of molecular superpixel representations. Besides computing summary features for each molecular superpixel as described in Section~\ref{sec:superpixel}, we learn patch representations via a MPN on the superpixel graph.
The ablation study were conducted on both tasks, ligand binding affinity (Section~\ref{sec:pdbbind}) and enzyme catalytic function classification (Section~\ref{sec:enzyme}).

As displayed in Table~\ref{tab:res_ablation}, \textsc{HoloProt} with (\coldiamond{darkblue}) and without molecular superpixels (\colcircle{lightblue}) improve over the performance of structure and surface representations. Further, the results of the ablation study clearly show that different protein scales are more relevant for particular downstream tasks, e.g., predicting the enzyme-catalyzed reaction class from surface only results in poor performance. We further see no improvement in applying a MPN within a molecular superpixel (\colsquare{pink}) over using summary features (\coldiamond{darkblue}). Further ablation studies are presented in Appendix~\ref{app:ablation}.

\subsection{Limitations} \label{sec:limitations}
\looseness -1 Despite the reported success of \textsc{HoloProt}, our method faces some limitations.
First, \textsc{HoloProt} relies on existing protein structures and the corresponding generated surface manifolds. However, protein sequence data still remains the most abundant data source, and in protein design, conformations of mutated macromolecules are unknown. This limitation could however be partly remedied, \begin{enumerate*}[label=(\roman*.)] 
\item by the recent advancements in protein structure prediction \citep[AlphaFold]{senior2020, jumper2021highly} \citep[RoseTTAFold]{baek2021accurate} and protein structure determination methods such as cryo-electron microscopy \citep{callaway2020a}, and
\item by utilizing homology modeling algorithms on available wild type structures for mutant analysis \citep{schymkowitz2005}
\end{enumerate*}. Second, our method requires precomputed surface meshes, resulting in an additional preprocessing step before deploying \textsc{HoloProt} to the desired application.
This bottleneck can be bypassed by utilizing techniques developed in the concurrent work by \citet{sverrisson2020}, which allow computation and sampling of the molecular surface on-the-fly.



\section{Conclusion} \label{sec:conclusion}
In this work, we present a novel multi-scale protein graph construction, \textsc{HoloProt}, which integrates finer and coarser representation details of a protein by connecting sequence and structure with surface. We further establish \emph{molecular superpixels}, which capture higher-level fingerprint motifs on the protein surface, improving the memory efficiency of our construction without reducing the overall performance. We validate \textsc{HoloProt}'s effectiveness and versatility through representative tasks on protein-ligand binding affinity and enzyme-catalyzed reaction class prediction.
While being significantly more parameter-efficient, \textsc{HoloProt} performs consistently well across different tasks and dataset splits, partly outperforming current state-of-the-art methods. This will potentially be of great benefit and advantage when working with datasets of reduced size, e.g., comprising experiments on mutational fitness of proteins, thus opening up new possibilities within protein engineering and design, which we leave for future work.

\section*{Acknowledgments}
This project received funding from the Swiss National Science Foundation under the National Center of Competence in Research (NCCR) Catalysis under grant agreement 51NF40 180544. Moreover, we thank Mojm\'ir Mutn\'y and Clemens Isert for their valuable feedback.

\bibliographystyle{abbrvnat}
\bibliography{main}

\newpage

\newpage

\section*{Appendix}

\appendix

\section{Message Passing Network} \label{app:mpn}
We utilize the Weisfeiler-Lehmann network (WLN) proposed in \citep{lei2017} as our base message passing network. This network builds a neural equivalent of the Weisfeiler-Lehmann test for comparing graphs. For clarity, we describe the network here. Consider a graph $\mathcal{G} = (\mathcal{V}, \mathcal{E})$. Given a node $v \in \mathcal{G}$ with neighbors $N(v)$, node features $\mathbf{f}_v$ and edge features $\mathbf{f}_{uv}$ for edge $(v, u) \in \mathcal{E}$, the WLN message passing step follows as,

\begin{equation}
\label{app:mpn_eq}
    \mathbf{m}_{v}^{(l)} = \tau(\mathbf{U_1m}_{v}^{(l-1)} + \mathbf{U_2}\sum_{u \in N(v)} \tau(\mathbf{V}[\mathbf{f}_u, \mathbf{f}_v])) \hspace{10pt} (1 \leq l \leq L)
\end{equation}

where $\tau(\cdot)$ could be any non-linear function, and $L$ is the total number of message passing steps, $\mathbf{h}_{v}^{(0)} = \mathbf{f}_v$.  The final representations for each node arise from
mimicking the set comparison function in the WL isomorphism test, yielding

\begin{equation}
    \mathbf{h}_v = \sum_{u \in N(v)}\mathbf{W}^{(0)}\mathbf{m}_u^{(L)} \odot
    \mathbf{W}^{(1)}\mathbf{f}_{uv} \odot \mathbf{W}^{(2)}\mathbf{m}_v^{(L)}
\end{equation}

\section{Multi-Scale Protein Representations}

\subsection{Ablation Studies} \label{app:ablation}

Table~\ref{tab:app_ablation} shows ablation study results for the \emph{identity 60\%} and \emph{scaffold} splits for \textsc{PDBBind} dataset. The table will be updated with values of \textsc{HoloProt} (molecular superpixels with MPN) setting once the results for the same are available.

\begin{table}[H]
\caption[Ablation Studies]{\textbf{Results of the Ablation Studies} Evaluation of architectural design choices of \textsc{HoloProt} by analyzing the performance of its individual components as well as feature summarization of molecular superpixels.}
    \label{tab:app_ablation}
    \begin{adjustbox}{max width=1\linewidth}
    \begin{tabular}{lccccccc}
    \toprule
    & \multicolumn{7}{c}{\textbf{Ligand Binding Affinity}} \\
    \textbf{Model} & \multicolumn{3}{c}{Sequence Identity (60 \%)} & &  \multicolumn{3}{c}{Scaffold} \\ \\
    & RMSE & Pearson & Spearman & & RMSE & Pearson & Spearman\\
    \cmidrule{2-4}\cmidrule{6-8}
    
    Structure & 1.378 $\pm$ 0.027 & 0.738 $\pm$ 0.014 & 0.730 $\pm$ 0.009 && 1.521 $\pm$ 0.023 & 0.485 $\pm$ 0.015 & 0.492 $\pm$ 0.013 \\
    Surface & 1.418 $\pm$ 0.014 & 0.719 $\pm$ 0.005 & 0.714 $\pm$ 0.004 && 1.558 $\pm$ 0.125 & 0.428 $\pm$ 0.159 & 0.429 $\pm$ 0.181 \\
    \textsc{HoloProt} (\colcircle{lightblue}) & \textbf{1.365} $\pm$ \textbf{0.038} & \textbf{0.749} $\pm$ \textbf{0.014} & \textbf{0.742} $\pm$ \textbf{0.011} && 1.523 $\pm$ 0.028 & 0.489 $\pm$ 0.019 & 0.491 $\pm$ 0.021 \\
    \textsc{HoloProt} (\coldiamond{darkblue}) & 1.473 $\pm$ 0.024 & 0.667 $\pm$ 0.011 & 0.675 $\pm$ 0.019 && \textbf{1.517} $\pm$ \textbf{0.014} & \textbf{0.491} $\pm$ \textbf{0.016} & \textbf{0.493} $\pm$ \textbf{0.014} \\
    \bottomrule \\
    \multicolumn{7}{c}{\large \colcircle{lightblue} \enspace full surface  \hfill \coldiamond{darkblue} \enspace molecular superpixels} 
    \end{tabular}
    \end{adjustbox}
\end{table}

\section{Superpixels on Molecular Surfaces}

\subsection{Computing Molecular Surfaces} \label{app:comp_surface}

Shape and surface of proteins determines their molecular interactions and thus, accurate computation of macromolecular surfaces from the provided atom point clouds is essential for elucidating their biological roles in physiological processes.
A variety of methods have been proposed to compute macromolecular surfaces. \emph{Van der Waals surfaces} is the simplest surface constructed via the topological
boundary of the set of atom spheres, each of van der Waals radius of the constituent atom. However, as most of the van der Waals surface is buried in the interior of large molecules, \citet{lee1971} defined the solvent-accessible surface (SAS), determined by the area traced out by the center of a probe sphere as it is rolled over the van der Waals surface. \citet{greer1978} proposed smooth solvent-excluded surfaces (SES, or \emph{molecular surface}) of a molecule \citep{connolly1983} defined as the boundary of the union of all possible probes having no intersection with the molecule. In this work, we utilize existing algorithm \texttt{MSMS} (Michel Sanner's Molecular Surface) computing triangulated representations of the \emph{molecular surface} relying on a reduced surface \citep{sanner1996}.

\paragraph{Details on Surface Preparation} 
All proteins were triangulated using the MSMS with a hydrogen density of $3.0$ and a water probe radius of $1.5$. The meshes are downsampled using BLENDER \citep{blender} to a uniform size of roughly $2600$ faces. In practice, we found that this size provided an appropriate balance between maintaining detail and memory consumption during preprocessing. Geometric and chemical features were computed directly on the protein mesh.

\subsection{Examples of Molecular Superpixels} \label{app:ex_superpixel}

\begin{figure}[h]
    \centering
    \includegraphics[width=.99\textwidth]{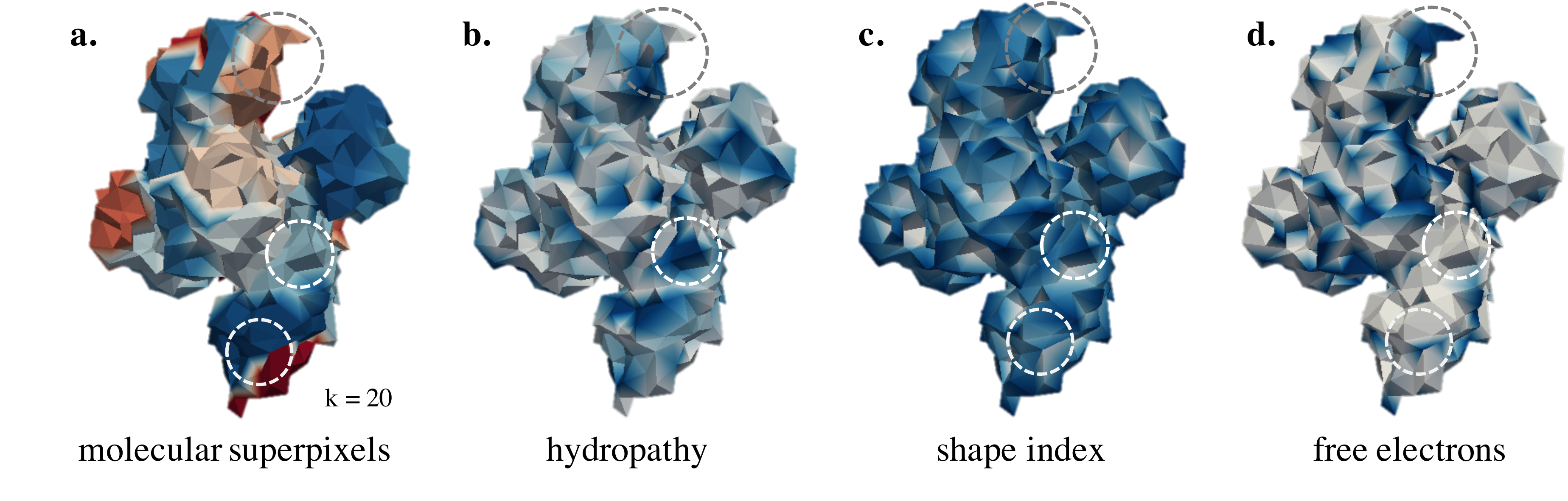}
    \caption{\textbf{Molecular Superpixels and Surface Features of the Hepatitis C Virus Helicase Inhibitor (\texttt{PDB ID: 4OKS}).} Molecular superpixels, indicated by different colors ($k = 20$), and the corresponding surface features, i.e., \textbf{b.} hydropathy, \textbf{c.}  shape index, and \textbf{d.} free electrons. As highlighted, molecular superpixels are spatially compact and overlap with surface regions dominated by single features such as hydrophobic patches while capturing coherent areas across all surface features. Protein complex contains 867 residues.}
    \label{fig:superpixel_4oks}
\end{figure}

\begin{figure}
    \centering
    \includegraphics[width=.99\textwidth]{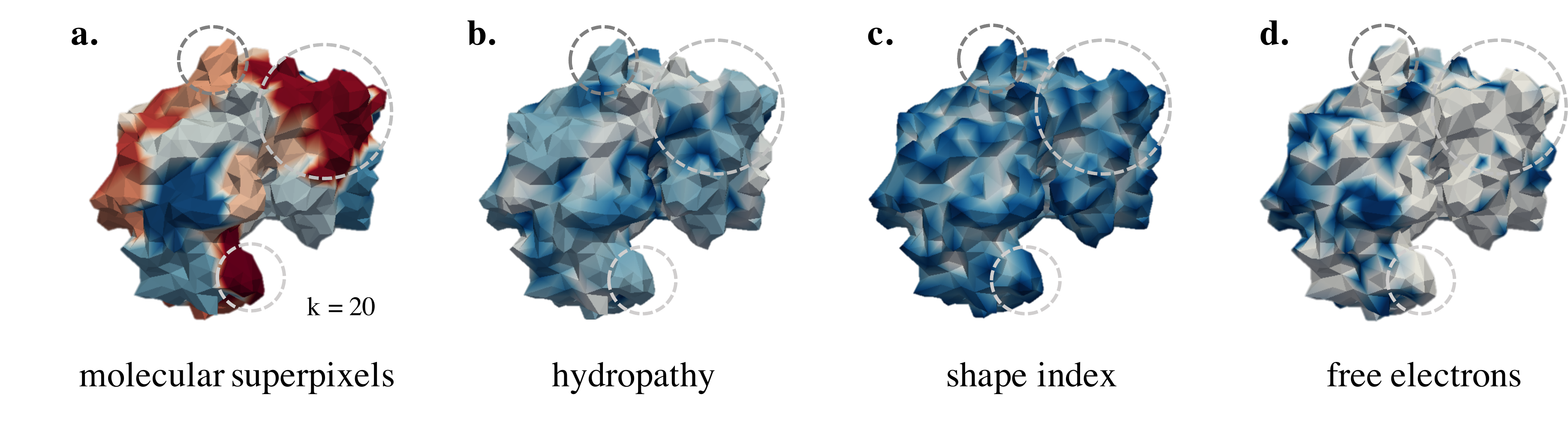}
    \caption{\textbf{Molecular Superpixels and Surface Features of Endothia Aspartic Proteinase (\texttt{PDB ID: 1EPO}).}  Molecular superpixels, indicated by different colors ($k = 20$), and the corresponding surface features, i.e., \textbf{b.} hydropathy, \textbf{c.}  shape index, and \textbf{d.} free electrons. As highlighted, molecular superpixels are spatially compact and overlap with surface regions dominated by single features such as hydrophobic patches while capturing coherent areas across all surface features. Protein complex contains 330 residues.}
    \label{fig:superpixel_1epo}
\end{figure}

Additional examples of molecular superpixels and their overlap with different surface features are shown in Figure~\ref{fig:superpixel_4oks} and Figure~\ref{fig:superpixel_1epo}.

\section{Experimental Details} \label{app:exp_details}
Our model is implemented in \texttt{PyTorch} \citep{paszke2019} using the \texttt{PyTorch Geometric} library \citep{fey2019}. We use the open-source software \texttt{RDKit} \citep{landrum2016}. We used \texttt{W\&B} \citep{biewald2020} for experiment tracking.

\subsection{Features} \label{app:features}

\subsubsection{Surface Layer}
We represent the surface layer as a graph $\mathcal{G}_{\mathcal{S}}$ where, for each node $u_\mathcal{S}$, we compute $4$ geometric and chemical features -- shape index, free electrons and proton donors, hydropathy and poisson-boltzmann electrostatics. These features are computed using code from \citep{gainza2020}, and the binaries APBS \citep{baker2001electrostatics}, PDB2PQR \citep{dolinsky2007pdb2pqr} and multivalue (provided within the APBS suite). We refer to \citep{gainza2020} for more details.

Two nodes share an edge if they are part of the same triangulation, and an edge is a part of two triangular faces. We compute $7$ edge features -- the dihedral angle between the two faces, the inner angles (one for each face) opposite to the edge, two edge-length ratios, where the edge ratio is between the
length of the edge and the perpendicular (dotted) line for each adjacent face. These features were taken from \citep{hanocka2019meshcnn}. We also include the distance between the surface nodes comprising the edge and the angle between the normals at those nodes.

\subsubsection{Structure Layer}
We represent the structure layer as a graph $\mathcal{G}_{\mathcal{B}}$ where the nodes $u_\mathcal{B}$ are the amino acid residues, and the edges occur between two amino acids within a certain distance threshold. We use the following node and edge features,

\begin{tabular}{l|c|c|c}
 \centering
 \textbf{Node Feature} & \textbf{Count} & \textbf{One-Hot} & \textbf{Possible Values}\\
 \toprule
 Residue Name & 23 & Yes & ALA, GLY etc. \\  
 Secondary structure the residue is part of & 8 & Yes & H, G, I, E, B, T, C, unk\\
 Solvent Accessible Surface of the residue & 1 & No & - \\
 Residue hydrophobicity & 1& No & - \\
\end{tabular}

As edge features, we use the angle between two residues and the distance between their  $\mathrm{C}_\alpha$ atoms. To compute the secondary structure, we use the DSSP binary \citep{kabsch1983dictionary}.

\subsubsection{Ligand Molecules}

We represent the ligand molecule as a graph $\mathcal{G}_\mathcal{L}$ with the following node and edge features,

\begin{tabular}{l|c|c|c}
 \centering
 \textbf{Node Feature} & \textbf{Count} & \textbf{One-Hot} & \textbf{Possible Values}\\
 \toprule
 Atom symbol & 65 & Yes & C, N, O etc.\\  
 Atom degree & 10 & Yes & 0, 1, 2, 3, 4, 5, 6, 7, 8, 9\\
 Implicit valence of the atom & 6 & Yes & 0, 1, 2, 3, 4, 5 \\
 Explicit valence of the atom & 6 & Yes & 1, 2, 3, 4, 5, 6\\
 Part of an aromatic ring & 1 & No & 0, 1 \\
\end{tabular}

\begin{tabular}{l|c|c|c}
 \textbf{Edge Feature} & \textbf{Count} & \textbf{One-Hot} & \textbf{Possible Values}\\
 \toprule
 Bond type & 4 & Yes & Single, Double, Triple, Aromatic \\
 Whether bond is conjugated & 1 & No & 0, 1 \\
 Whether bond is part of ring & 1 & No & 0, 1 \\
\end{tabular}

\subsection{Datasets} \label{app:datasets}

\paragraph{Protein-Ligand Binding Affinity} We use the refined subset of the 2019 version of PDBBind \citep{liu2017} and evaluate our model on 3 splits -- \emph{scaffold}, \emph{identity 60\%} and \emph{identity 30\%}. Two of these splits (\emph{identity 60\%} and \emph{identity 30\%}) are based on sequence identity, with sequences in the test set not having more than a 30\% or 60\% to sequences in the training set. We use the same splits as provided by \citep{townshend2020}, and refer the reader to the same for more details on their construction. The scaffold split is prepared by computing the Bemis-Murcko scaffold \citep{bemis1996properties} using \texttt{RDKit} for each molecule, and splitting the molecules such that molecules with rare or unseen scaffolds are part of the test set.

\paragraph{Enzyme-Catalyzed Reaction Classification} We use the PDB files and splits provided by \citet{hermosilla2021} for this task. For more details on the dataset construction, we refer the reader to \citet[\S~C]{hermosilla2021}.

\subsection{Network Architectures} \label{app:hyperparams}

\paragraph{Hyperparameter Tuning}
For protein-ligand binding affinity prediction, we performed a hyperparameter sweep over the hidden dimensions of the surface ($150, 200, 300$) and structure layers ($150, 200, 300$), and the hidden dimensions of the MLP ($512, 256$, [$512, 256$]). For the enzyme-catalyzed reaction classification, given time constraints, our hyperparameter tuning was restricted to the learning rates $0.001, 0.0005, 0.0001$ and hidden layer activations ($\mathrm{ReLU}, \mathrm{LeakyReLU}$).

For all \textsc{HoloProt} models, we use the Adam optimizer for training and clip gradients to a maximum norm of $10.0$.

\subsubsection{Protein-Ligand Binding Affinity}
For \textsc{HoloProt} with the full surface, the surface and structure layer MPNs have hidden dimensions of $150$ and $200$ with message passing steps of $6$ and $5$ respectively. The affinity prediction MLP has a single hidden layer of dimension $512$. For \textsc{HoloProt} with molecular superpixels, the surface and structure layer MPNs have hidden dimensions of $150$ and $300$, with $5$ message passing steps. The affinity prediction MLP has a single hidden layer of dimension $256$. For both models, the ligand MPN has a hidden layer dimension of $300$, with $4$ message passing steps. We use the $\mathrm{ReLU}$ activation function. Starting with an initial learning rate of $0.001$, we apply a learning rate decay of $0.9$ based on a validation RMSE plateau, with an improvement threshold of $0.01$ and a patience of $5$. The \textsc{HoloProt} model for full surface has $1.44$M parameters, while the \textsc{HoloProt} model with molecular superpixels has $1.76$M parameter.

\subsubsection{Enzyme-Catalyzed Reaction Classification}
Both \textsc{HoloProt} models have a hidden dimension of $150$ for both the surface and structure layers, with $4$ and $5$ message passing steps. The classification MLP has a single hidden layer of dimension $512$. We use the $\mathrm{LeakyReLU}$ activation function, and apply dropout with probability $0.15$ for each message passing step, and $0.3$ for the classification MLP. Starting with an initial learning rate of $0.0005$, we apply a learning rate decay of $0.6$ based on a validation accuracy plateau, with an improvement threshold of $0.01$ and a patience of $10$. Both models have roughly $0.64$M parameters.
We ran \textsc{HoloProt} with more parameters (5.2 M) \emph{without} hyperparameter search due to computational restrictions, resulting in small improvement of the overall performance (i.e., an accuracy of $79.2 \%$ on the enzyme-catalyzed reaction classification task).

\subsubsection{Baselines}
For protein-ligand binding affinity prediction, we use the provided code for different baselines and extend them as necessary for the task. For enzyme-catalyzed reaction classification, we use the baseline values from \citep{hermosilla2021}. 

Across all models for protein-ligand binding affinity prediction, we compute representations for proteins and ligands and concatenate them and use that as input for the MLP. For the models described in \citep{townshend2020}, we use the reported values. Wherever possible, we restrict the ligand embedding dimension to be $300$ consistent with our experiments. The details for the remaining baselines are as follows,

\paragraph{\citet{ozturk2018}} We downloaded the code from the official repository. The authors do not use a separate validation set, but instead use a cross-validation strategy. We combine the training and validation sets and then perform 5-fold cross-validation. The authors allow specification of hyperparameters for the number of filters ($32$, $64$), the size of ligand sequence filters ($4$, $8$, $12$, $36$), and size of protein sequence filters ($4$, $8$, $12$, $36$). We use a default batch size of $64$, and the default learning rate, and train the model for $100$ epochs. We also note that the model typically undergoes early stopping around epoch $80$. The best performing model occurred with $32$ filters, with a filter size of $4$ for ligand sequences, and $8$ for protein sequences.

\paragraph{\citet{bepler2019}} We downloaded the code and pretrained models from the official repository and embed the ligand sequence with a bidirectional LSTM. For embedding the protein, we use the default values as from the pretrained model. For the ligand, we use an input dimension of $512$, and a LSTM hidden dimension of $512$, with a final embedding dimension of $100$, similar to the protein. Our hyperparameter tuning is restricted to ligand input dimensions ($512, 256$), LSTM hidden dimensions ($512 , 256$) and the MLP hidden dimensions ($512, 256$, [$512, 256$]). We use a batch size of $32$ and learning rate of $0.00001$ for the pretrained model, and $0.001$ for the remaining parameters.

\paragraph{\citet{rao2019}} We downloaded the code and pretrained models from the official repository. We represent the ligand as a sequence and embed it using a Transformer, with an embedding dimension of $300$ and an intermediate size of $512$. We use a 2-layer MLP with hidden dimensions of $512$ and $256$ and dropout probabilities of $0.2$ to predict the binding affinity after concatenating the protein and ligand embeddings. For training, we use a learning rate of $0.0001$ for the pretrained model parameters and $0.001$ for the remaining parameters, and trained the model for $300$ epochs. Our hyperparameter sweep was restricted to batch size (default value $32$, multiplied and divided by $2$ until no improvement), and hidden layer dimensions for the MLP ($512$, $256$, [$512$, $256$]).

\paragraph{\citet{gainza2020}} We downloaded the code from the official repository and extended the model for the ligand binding affinity task. For the protein, we use the default values provided. We represent the ligand as a graph and use the same architecture and parameters as our message passing network, with a hidden dimension of $300$. Given memory and time constraints, we were unable to perform a hyperparameter sweep.

\paragraph{\citet{hermosilla2021}} We downloaded the code from their official repository. The proteins are embedded with an embedding dimension of $1024$. We represent the ligand as a graph and use the same architecture and parameters as our message passing network, with a hidden dimension of $300$. We concatenate the protein and ligand embeddings before using it as input for a single-layer MLP with hidden size $512$. The model is trained with the default learning rate $0.001$ and learning rate decay for $300$ epochs. Due to memory constraints, we trained with the default batch size of $8$, and performed a hyperparameter sweep for MLP hidden sizes $512$, $256$, [$512, 256$], and protein embedding dimensions $1024, 512$. 

\subsection{Computing Infrastructure} \label{app:computing}
All models were trained on a single NVIDIA 1080Ti GPU. For the PDBBind dataset, all the \textsc{HoloProt} models run within $16$ hours when trained for $200$ epochs, and within $8$ hours when trained for $100$ epochs. The baseline models \citep{rao2019, hermosilla2021} take about $40$ hours for running $300$ epochs, while the remaining baseline models train in under $24$ hours. For the enzyme-catalyzed reaction classification dataset, the \textsc{HoloProt} models are trained for $24$ hours on the NVIDIA 1080Ti GPU after which it is stopped. Typically, the model is trained for $100$ to $110$ epochs by then, when using a batch size of $10$.


\end{document}